\documentclass[3p,twocolumn]{elsarticle}

\usepackage{amssymb}
\usepackage[fleqn]{amsmath}

\newenvironment{hangref}
  {\begin{list}{}{\setlength{\itemsep}{4pt}
  \setlength{\parsep}{0pt}\setlength{\leftmargin}{+\parindent}
  \setlength{\itemindent}{-\parindent}}}{\end{list}}

\newcommand{\xw}{8.0cm}

\begin{document}

\begin{frontmatter}


 \author{Gerald Paul }
 \ead {gerryp@bu.edu}

\title{An efficient  implementation of the simulated annealing heuristic for the quadratic assignment problem}

 \address{Center for Polymer Studies and Dept. of Physics, Boston University}
\address{Boston, Massachusetts 02215}



\begin{abstract} The quadratic assignment problem (QAP) is one of the
  most difficult combinatorial optimization problems.  One of the most
  powerful and commonly used heuristics to obtain approximations to
  the optimal solution of the QAP is simulated annealing (SA).  We
  present an efficient implementation of the SA heuristic which
  performs more than 100 times faster then existing implementations for
  large problem sizes and a large number of SA iterations.

\end{abstract}

\begin{keyword}


\end{keyword}

\end{frontmatter}


\section{Introduction}

Originally  formulated by Koopmans and Beckmann (1957), the QAP is NP-hard and is considered to be one of the most difficult problems to
be solved optimally. It was defined in the following context: A set of $N$ facilities are to be located at $N$ locations.
The quantity of materials which flow between facilities $i$ and $j$ is
$A_{ij}$ and the distance between locations $i$ and $j$ is $B_{ij}$.
The problem is to assign to each location a single facility so as to
minimize (or maximize) the cost
\begin{equation*}
C=\sum_{i=1}^N \sum_{j=1}^N  A_{ij} B_{p(i)p(j)},	
\end{equation*}
where $p(i)$ represents the location to which $i$ is assigned.

The QAP formulation has since been applied to such diverse problems as the optimal assignment of classes to classrooms (Dickey and Hopkins, 1972), design of
  DNA microarrays (de Carvalho and Rahmann, 2006), cross species gene analysis
  (Kolar, et al., 2008), creation of hospital layouts (Elshafei, 1977), and
  assignment of components to locations on circuit boards (Steinberg, 1961).

In addition to being important in its own right, the QAP includes such
other combinatorial optimization problems as the traveling salesman
problem and graph partitioning as special cases. 

There is an extensive literature that addresses the QAP and which is
reviewed in Pardalos et al. (1994), Cela (1998), Anstreicher (2003),
Loiola et al. (2007), James et al. (2009) and Burkhard et al. (2009). The QAP is exceedingly
hard to solve optimally.  With the exception of specially constructed cases, optimal
algorithms have solved only relatively small instances with $N \le
36$. Various heuristic approaches have been developed and applied to
problems typically of size $N\approx 100$ or less.  By contrast, a
travelling salesman problem consisting of almost 25,000 towns in
Sweden has been solved exactly (Applegate et al., 2007).

\section{Background}

The simulated annealing heuristic was first applied to the QAP by
Burkhard and Rendl (1984) and was refined by Connolly (1990).  The
heuristic consists of swapping locations of two facilities.  Proposed
swaps can either be determined randomly or selected according to some sequential
enumeration of all possible swaps.  For each proposed swap, $\delta$, the change in cost for the potential
swap, is calculated.  The swap is made if $\delta$ is negative or if
\begin{equation*}
e^{-\delta/T} > r,
\end{equation*}
where $T$ is an analog of temperature in physical
systems that is slowly decreased according to a specified {\it cooling
  schedule} after each iteration and $r$ is a uniformly distributed random variable between 0 and 1.  Randomly making moves
which increase the cost is done to help escape from local minima.

In traditional implementations of the heuristic, the cost of making a
swap is calculated from scratch when the swap is considered in order
to determine if the swap should be made.  The computational complexity
of this calculation is $O(N)$; and if the SA run is composed of $I$
iterations, the complexity is $O(I N)$.  We can write the time
needed to execute $I$ iterations as
\begin{equation}
t_{trad}=c_0 NI
\label{trad}
\end{equation}
where  $c_0$ is a constant.

\section{Approach}
\label{approach}

The approach we take here is to attempt to reduce the complexity by
maintaining a matrix $\Delta$ of costs of each possible swap.  This
approach is motivated by the work of Taillard (1991), who
applied a similar approach in the application of another heuristic,
tabu search, to the QAP.  Our $\Delta${\it -matrix  approach} is as follows:

\begin{itemize}

\item[(a)] Initialize by creating a matrix  $\Delta(p,u,v)$ containing the cost of swapping 
$u$ and $v$ for all $u$ and $v$, given a current assignment $p$. 
\item[(b)] In each iteration, retrieve the  cost   $\Delta(p,r,s)$ of the proposed swap $(r,s)$.
\item[(c)] If the proposed swap is not made in a given iteration, go to (b).
\item[(d)] If  the proposed swap is made in a given iteration, update $\Delta(p,u,v)$  to reflect the swap costs given the new permutation and go to (b).
\item [(e)] End after $I$ iterations.
\end{itemize}

The number of iterations in which a swap is performed divided by the total number of iterations is known as  the {\it acceptance rate} $a(I)$.

The computational complexity of our approach has the following contributions.

\begin{itemize}
\item[(i)]     Retrieving the value of $\Delta(p,r,s)$  is of complexity $O(1)$.
\item[(ii)]    Assuming an acceptance rate, $a(I)$, the matrix  $\Delta(p,r,s)$ must be updated  $a(I)I$ times.   The complexity of updating $\Delta(p,u,v)$ is  $O(N^2)$ as described in  \ref{complexity}.
\end{itemize}

Thus the complexity of our approach is 
\begin{equation*}
O(a(I)I N^2) + O(I)
\end{equation*}
and we can write the CPU time  of our approach $t_{\Delta}$ as
\begin{equation*}
t_{\Delta}=c_1 a(I)IN^2 + c_2 I
\end{equation*}
where $c_1$ and $c_2$ are constants.  The performance improvement factor  of our approach  relative to the traditional approach is then
\begin{equation}
  F \equiv  \frac{t_{trad}}{t_{\Delta}  } =       \frac{c_0 N}{c_1 a(I)N^2 + c_2} \to \frac{c_0}{c_1 a(I)N }
\label{F}
\end{equation}
for large $N$.

From Eq. (\ref{F}) we see that the performance improvement is
critically dependent on the acceptance rate $a(I)$.  The constants
$c_0$ and $c_1$ are independent of the problem instance, depending
only on the SA implementation; for our implementation $c_0/c_1 \approx
1/3$.  For our approach to be beneficial, $F$ must have a value $> 1$
and thus $a(I)$ must satisfy
\begin{equation}
 a(I)    \lesssim 1/3N.  
\label{critical}
\end{equation}
It is useful to consider an {\it instantaneous acceptance rate},
$\tilde a(i)$, defined as the number of swaps performed in a window of iterations around iteration $i$ divided by
the size of the window.  Our numerical experiments indicate that $\tilde a(i)$
decreases with increasing $i$ as the SA run progresses (see
\ref{accAPP}).  For this reason, our program uses the traditional
approach until an iteration at which $\tilde a(i)$ satisfies Eq.
(\ref{critical}) and then begins using the more efficient
$\Delta$-matrix approach.

For tabu search, a swap is made only after all possible $N(N-1)/2$
swaps are analyzed.  Thus for tabu search, the acceptance rate, $a
\sim 1/N^2$ and the performance improvement is of O(N).  However, for
simulated annealing $a$ is a function of the QAP instance and the
number of iterations performed in a run.  We know of no method of
determining the acceptance rate {\it a priori}.  Hence, in section
\ref{acceptanceRate} we measure $a(I)$ for different QAP instances and
various values of $I$ to get a sense of the performance improvement we can expect.

\section{Numerical Results}

We perform numerical experiments on two classes of QAP instances:
random instances and structured instances.  The random instances are
created in the same manner as the Taixxa instances (Taillard, 1991) in
QAPLIB (Burkard et al. 1997); the $A$ and $B$ matrices are symmetric
with zero diagonal; the matrix elements are chosen from independent
uniform distributions.  The structured instances are the Taixxxe
instances created by Taillard ( Drezner et al., 2005) designed to
model real-world applications which are difficult to treat using
iterative heuristics such as SA; the instances are composed of
matrices which are not random They are available at
http://mistic.heig-vd\ .ch/taillard/.

All runs were performed on systems with Intel Xeon 2.4 GHz processors.

\subsection{Acceptance Rate}
\label{acceptanceRate}

To get an idea of the performance improvement we may expect, we
measure the acceptance rate for a range of problem sizes and number of
iterations for the problem instances discussed above.  We use the c++
implementation written by Taillard that implements the SA heuristic of
Connolly(1990). Taillard's code is available at http://mistic.heig-vd\
.ch/taillard/ .

\subsection{Random instances}

In Fig. \ref{pacc}(a), we plot the acceptance rates $a(I)$ for the
random instances versus the number of iterations performed for various
problem sizes.  We note that $a(I)$ decreases with increasing $I$ until reaching a minimum after which $a(I)$ increases slowly.
The location of the minimum in $a(I)$ is dependent on $N$; the larger the problem size,
the higher the number of iterations at which the minimum is reached.
In \ref{accrApp}, we provide insights into this behavior.

We also note that for large numbers of iterations, $a(I)$ is lower for
larger problem sizes.  This is fortuitous because from Eq. (\ref{F}) a
lower value of $a(I)$ is needed to achieve the same performance
improvement factor if $N$ is increased.

\subsection{Structured instances}

In Fig. \ref{pacc}(b), we plot the acceptance rate for the structured
instances.  The overall behavior is the same as for the random
instances, but after the minimum is reached, $a(I)$ does not
increase with $I$.  In \ref{accsApp}, we provide explain this
behavior.  Also for instances of approximately the same size, $a(I)$ is
higher for the structured instances than for the random instances.
This implies that the performance improvement for the structured
instances will not be as great as for the random instances.

\subsection{Performance Improvement}

\begin{figure}
$
\begin{array}{c}

\includegraphics[width=\xw]{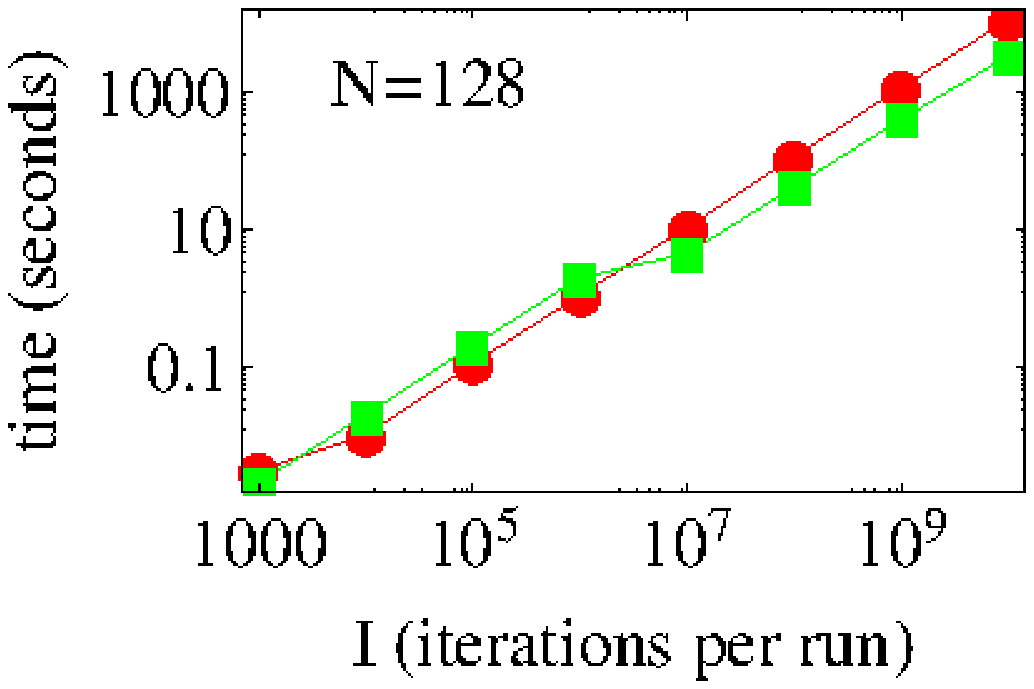}
\includegraphics[width=\xw]{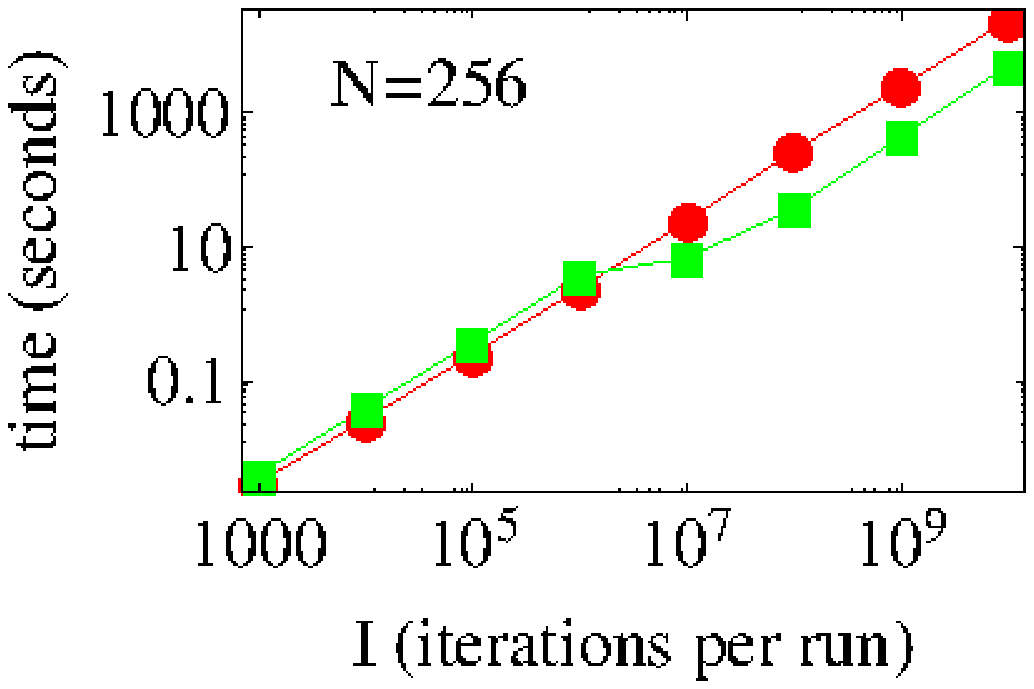} \\

\includegraphics[width=\xw]{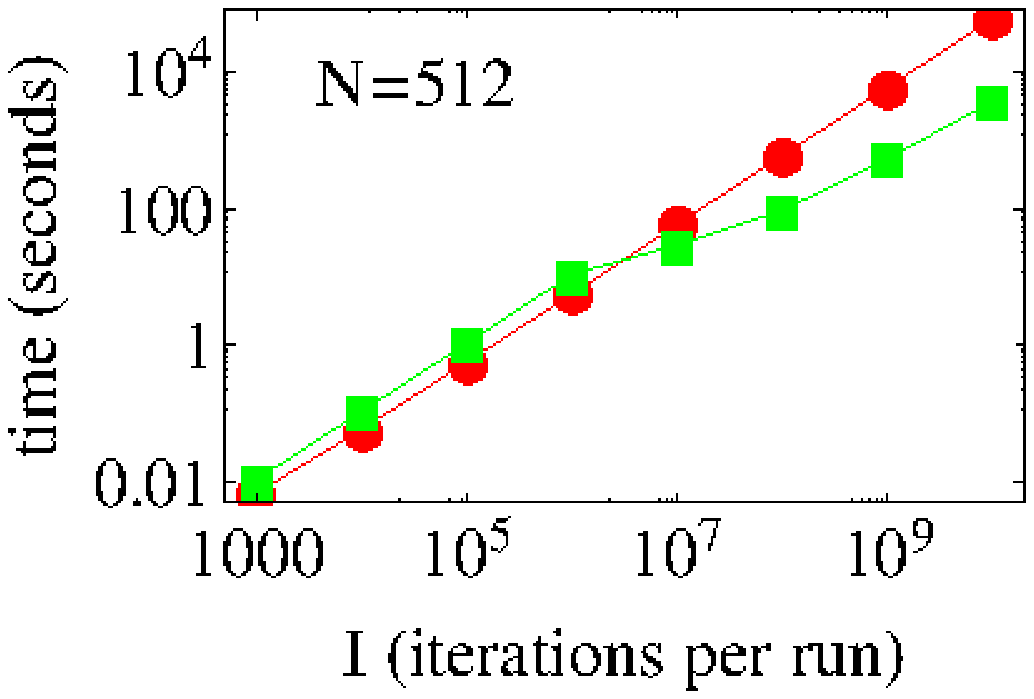}
\includegraphics[width=\xw]{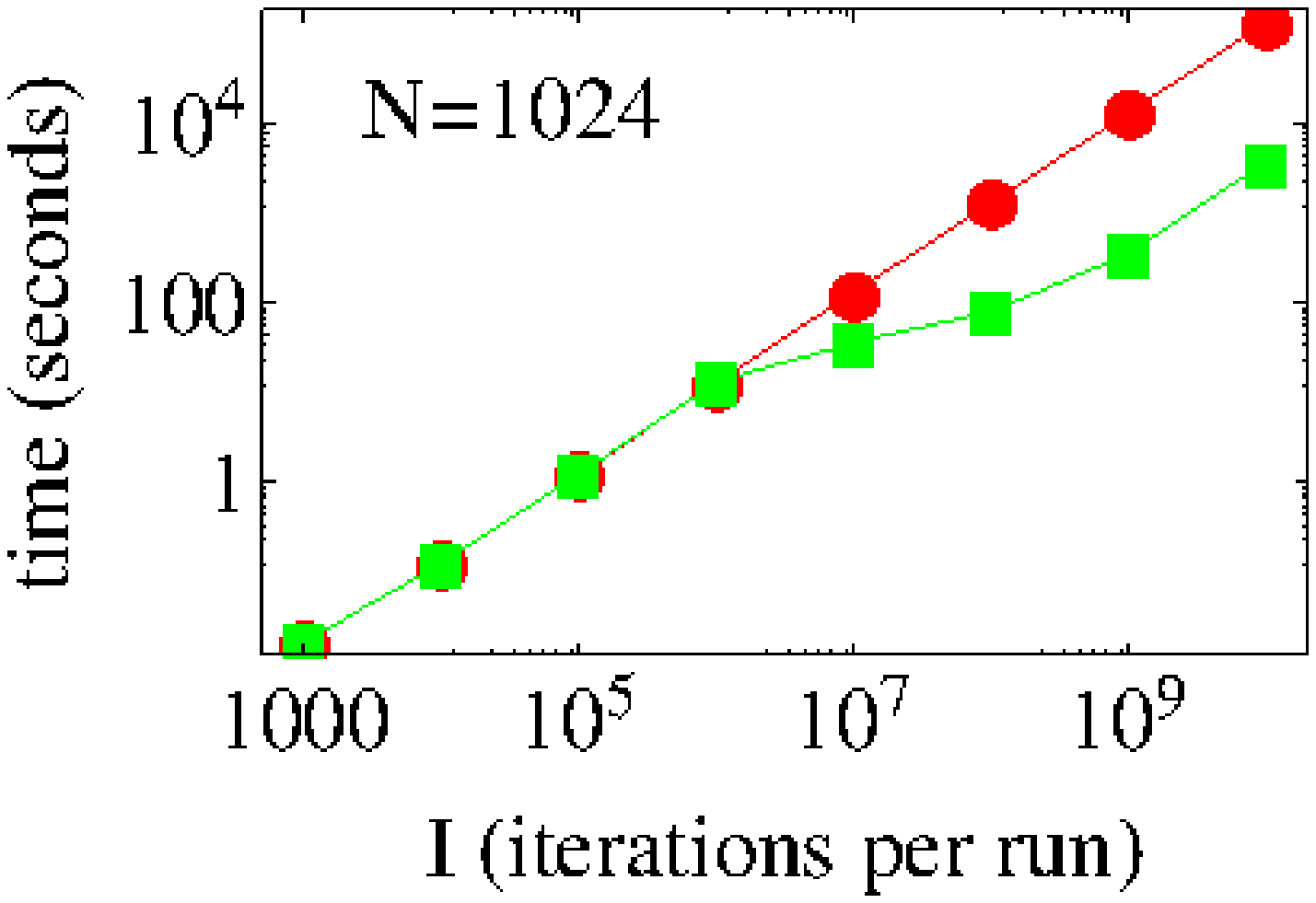} \\

\includegraphics[width=\xw]{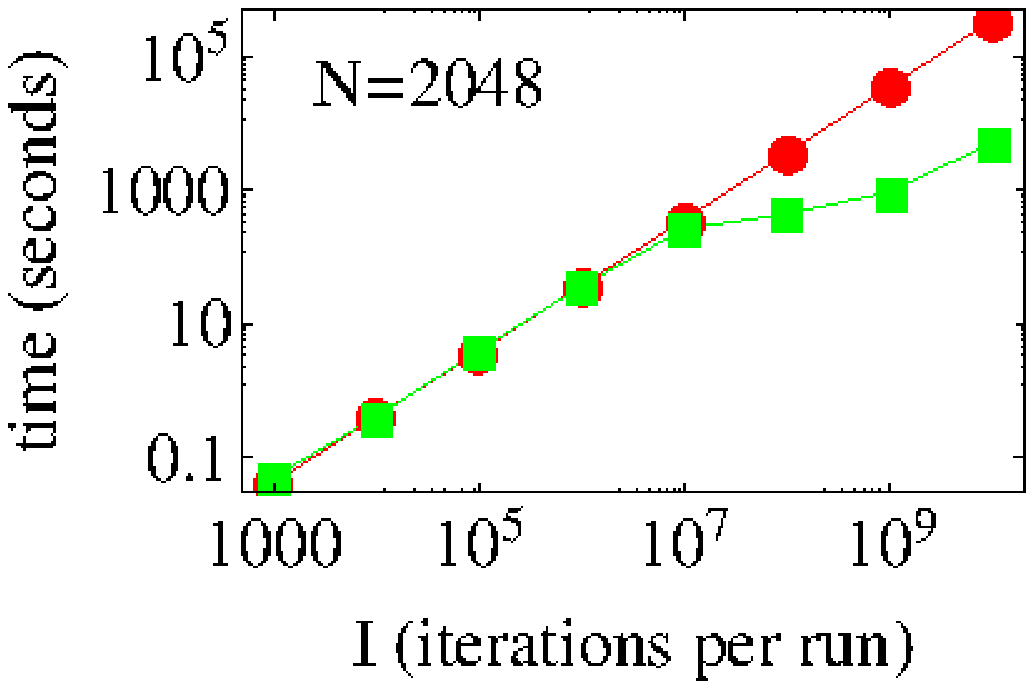}
\includegraphics[width=\xw]{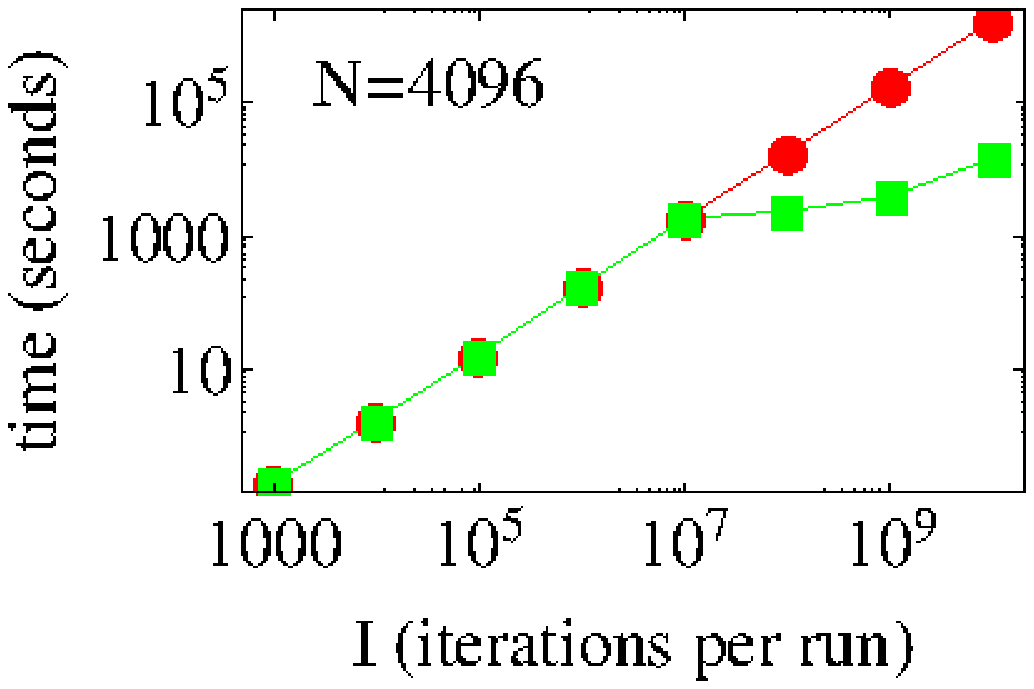} \\

\end{array}
$
\caption{CPU time versus iterations for random instances for traditional implementation (circles) and efficient implementation (squares). }
\label{pxxxr}
\end{figure}

\begin{figure}
\includegraphics[width=\xw]{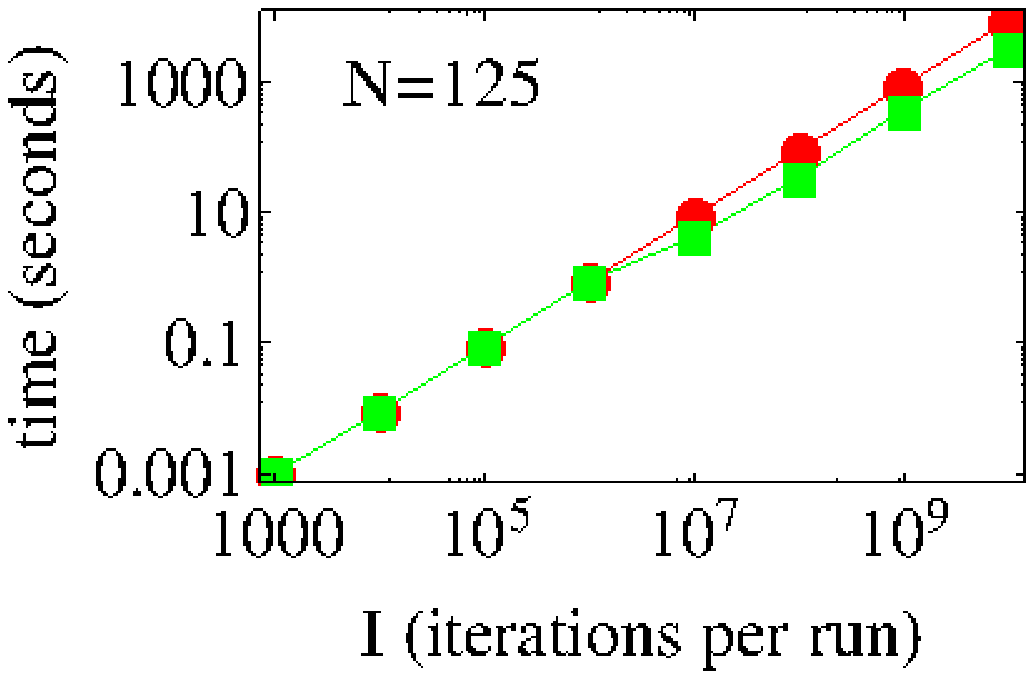} \\
\includegraphics[width=\xw]{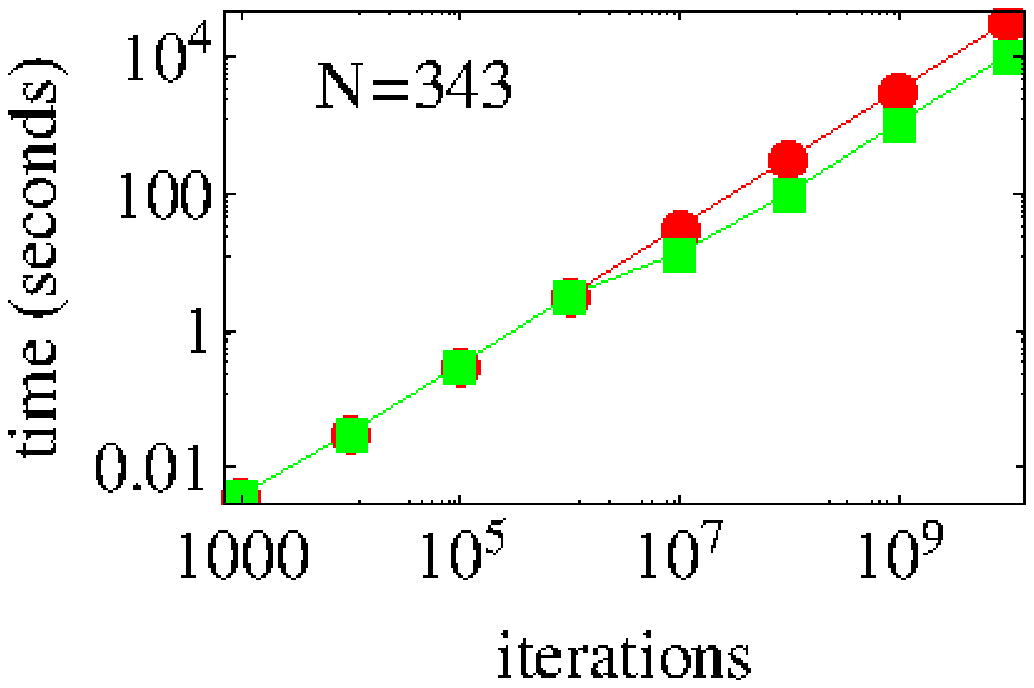} \\
\includegraphics[width=\xw]{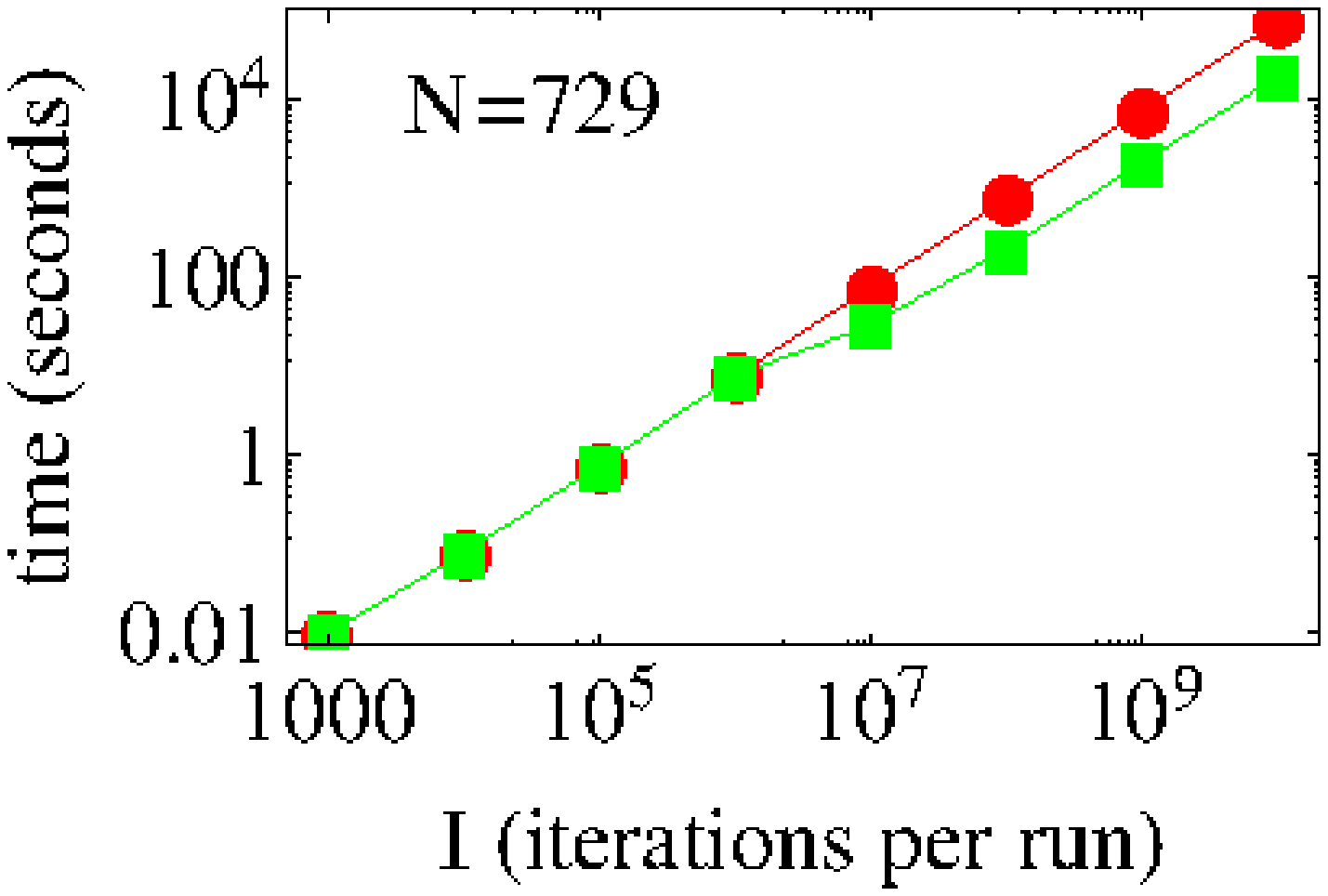}
\caption{CPU time versus iterations for structured  instances for traditional implementation (circles) and efficient implementation (squares). }
\label{pxxxs}
\end{figure}

We use the Taillard code as a base for our implementation described in
Section \ref{approach}.  For the random instances, in Fig. \ref{pxxxr}
we plot CPU time vs the number of iterations for the traditional
approach and for our $\Delta$-matrix approach.  Note that the times
for the traditional approach are linear with the number of iterations,
consistent with Eq. (\ref{trad}).  The $\Delta$-matrix approach
results are coincident with those of the traditional approach until
Eq. (\ref{critical}) is satisfied, at which point it is more efficient
to use the $\Delta$ matrix.  From this point on, the $\Delta$-matrix
approach CPU times are lower than those of the traditional approach.
Similar results hold for the structured instances as shown in Fig.
\ref{pxxxs}.

\begin{figure}
\includegraphics[width=\xw]{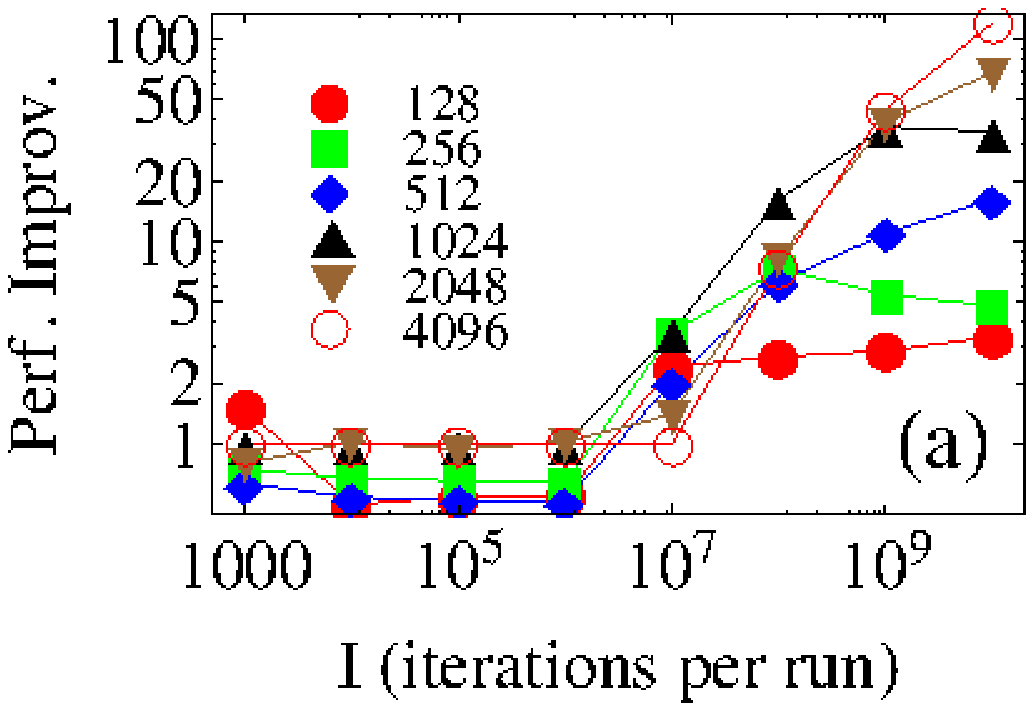}
\includegraphics[width=\xw]{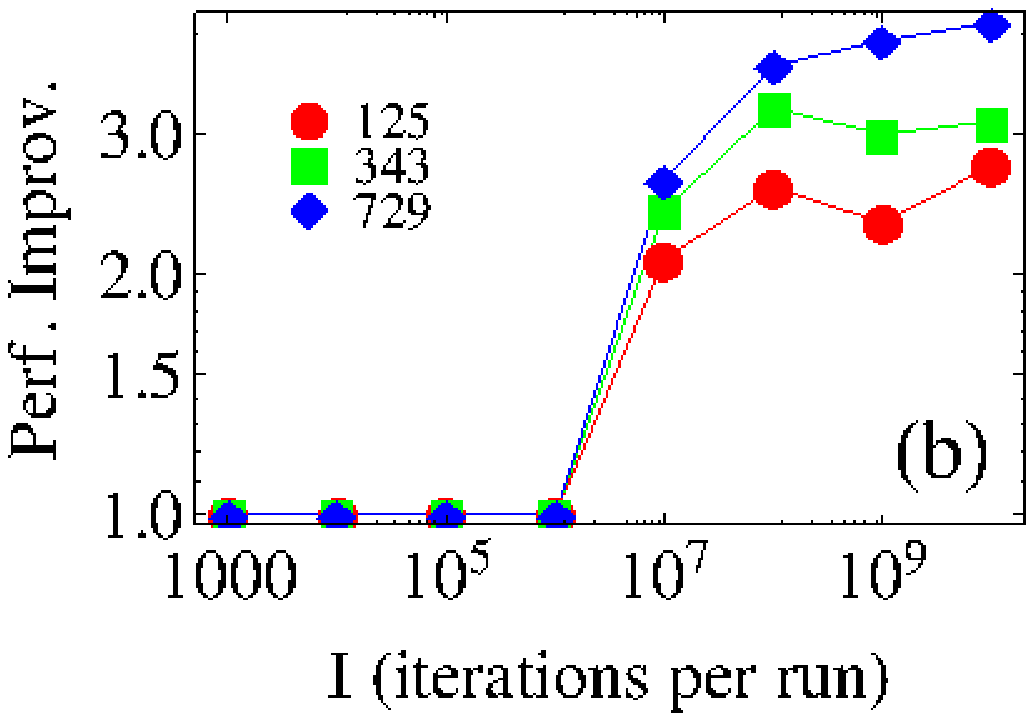}
\caption{Performance improvement for (a) random instances and (b) structured instances.  }
\label{ps}
\end{figure}

In Fig. \ref{ps}, we plot the measured performance improvement
$t_{trad}/t_{\Delta}$ for random and structured instances.  There is no
performance improvement until the number of iterations $\sim 10^6$.
This reflects the fact that below this number of iterations Eq.
(\ref{critical}) does not hold and the traditional method is used.  We
see that the relative performance of the $\Delta$-matrix approach
improves with increasing problem size, from which we can infer from Eq.
(\ref{F}) that the acceptance rate decreases faster than the problem
size as the problem size increases.

\section{Discussion and Summary}

We develop an implementation of the simulated annealing heuristic for
the quadratic assignment problem which runs significantly faster than
the traditional implementation; the simulated annealing concept and
algorithm for the QAP are unchanged.  Our approach is motivated by the
work of Taillard (1991) who took a similar approach in
implementing tabu search for the quadratic assignment problem.  That
the technique has not been applied to SA until now may be due to the
fact that it does not provide significant benefit unless a large
number of iterations are applied to a large QAP instance; this is now
possible with the improvement of computing capabilities over the last
20 years.

Being able to perform SA with a large number of iterations on large
QAP instances is important because (a) many real world problems are
significantly larger than the size of problems to which SA has been
traditionally applied and (b) the quality of the SA solution improves
as the number of iterations is increased (Paul, 2010).  Also, in Paul (2010) it was shown that for high quality solutions, the
performance of SA is superior to that of tabu search.  The current work makes
this finding stronger because of the improved performance of SA.

\begin{figure}
\includegraphics[width=\xw]{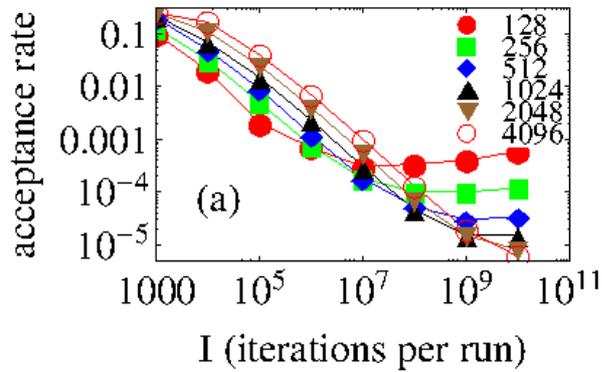}
\includegraphics[width=\xw]{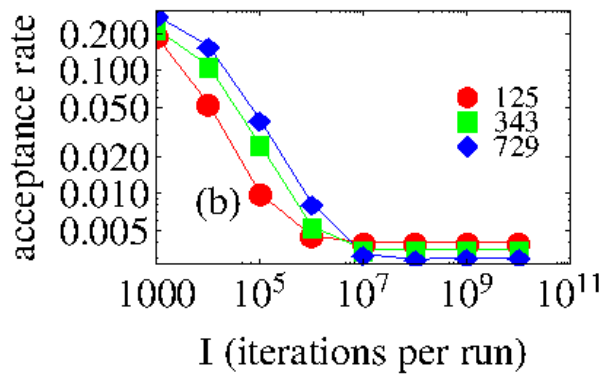}
\caption{Acceptance rates for(a) random and (b) structured instances versus number of iterations.  }
\label{pacc}
\end{figure}

\newpage
\twocolumn

\onecolumn

\twocolumn

 \appendix
\setcounter{figure}{0}

\section{Complexity of $\Delta$-Matrix Update}
 \label{complexity}

 Following Taillard (1991), starting from an assignment of facilities
  $p$ let the resulting assignment after swapping facilities $r$ and
  $s$ be $p'$.  That is:
\begin{eqnarray*}
p'(k) &=& p(k) ~~~~~   k\ne r,s               \nonumber \\
p'(r) &=& p(s)                                \nonumber \\
p'(s) &=& p(r).  
\end{eqnarray*}
For a symmetrical matrix with a null-diagonal, the cost
$\Delta(p,r,s)$ of swapping $r$ and $s$ is:
\begin{eqnarray}
  \Delta(p,r,s) &=&  \sum_{i=1}^N \sum_{j=1}^N (A_{i,j} B_{p(i),p(j)} - A_{i,j} B_{p'(i),p'(j)}) \nonumber \\
  &=& 2 \sum_{k \ne r,s} (A_{s,k} -A_{r,k}) (B_{p(s),p(k)-B_{p(r)},p(k)}).                        \nonumber \\
\label{eqfull}
\end{eqnarray}

To calculate $\Delta(p',u,v)$ for any $u$ and $v$ with complexity O(N)
, we can use equation (\ref{eqfull}). For asymmetric matrices or matrices with
non-null diagonals, a slightly more complicated version of Equation
(\ref{eqfull}) also of complexity O(N) is given by Burkhard and Rendl (1984).

In the case that $u$ and $v$ are different from $r$ or $s$,  we can  calculate $\Delta(p',u,v)$ with complexity $O(1)$ with
\begin{eqnarray*}
\Delta(p',u,v) = \Delta(p,u,v)                                 
+2(A_{r,u} - A_{r,v} + A_{s,v} -A_{s,u})                \nonumber \\
\times (B_{p'(s),p'(u)} - B_{p'(s),p'(v)} + B_{p'(r),p'(v)} - B_{p'(r),p'(u)})
\end{eqnarray*}
using the value $\Delta(p,u,v)$ calculated in the previous iteration.  Since there are $O(N^2)$ matrix entries with  $u$ and $v$ different from $r$ or $s$, these contribute complexity $O(N^2)$ to the $\Delta$ matrix update.  There are $O(N)$ entries with  $u$ and $v$ not different from $r$ or $s$ each of which requires the $O(N)$ calculation of Eq. (\ref{eqfull}). These entries thus also contribute complexity $O(N^2)$ and the overall complexity of  updating $\Delta$ is $O(N^2)$.

\section{Acceptance rate behavior }
 \label{accAPP}

 Because the performance improvement depends critically on the
 acceptance rate as shown in Eq. (\ref{F}), here we explore the
 behavior of the acceptance rate as shown in Fig. \ref{pacc}.  In
 order to do this, we first provide more details of the cooling
 schedule of the SA algorithm (Connolly, 1990) and describe how an SA
 run progresses.

 The details of the cooling schedule are as follows: At the start of
 each heuristic run, the temperature is set to a high value $T_0$.
 After each iteration, the temperature is reduced using the recursive
 relation
\begin{equation}
T=\frac{T}{1.0 + \beta T}
\end{equation}
where
\begin{equation}
\beta \equiv \frac{T_0 - T_f} {I T_0 T_F}.
\label{beta}
\end{equation}
to reach a desired final temperature $T_f$ when all iterations are
exhausted.  Note from Eq. (\ref{beta}) that the temperature decreases
in smaller steps for larger $I$.  The temperature decrease may end
before $T_f$ is reached if a specified number, $Q$, of consecutive
iterations in which no swap is performed is encountered. We denote the
temperature at which this occurs as $T_Q$.

At the beginning of an SA run, the initial QAP
configuration is random and there are relatively many moves which will
result in a cost decrease.  As the run progresses, the number of moves
which will result in a lower cost decreases.  Also as the run
progresses, the decreasing temperature results in a lower frequency of
swaps with a positive incremental cost $\delta$.  Thus the instantaneous acceptance
rate $\tilde a(i)$  (defined in Section \ref{approach}) decreases as the run progresses.  Eventually if the number of
iterations $I$ specified for the run is large enough, the number of
consecutive iterations in which a swap is not performed will exceed
the specified limit $Q$ and the temperature and $\tilde a(i)$
will no longer decrease.

With this background we now discuss the behavior of the acceptance
rates $a(I)$  shown in Fig.  \ref{pacc} for both the random and structured
instances.

\subsection{Random instances}
\label{accrApp}

\begin{figure}
\includegraphics[width=\xw]{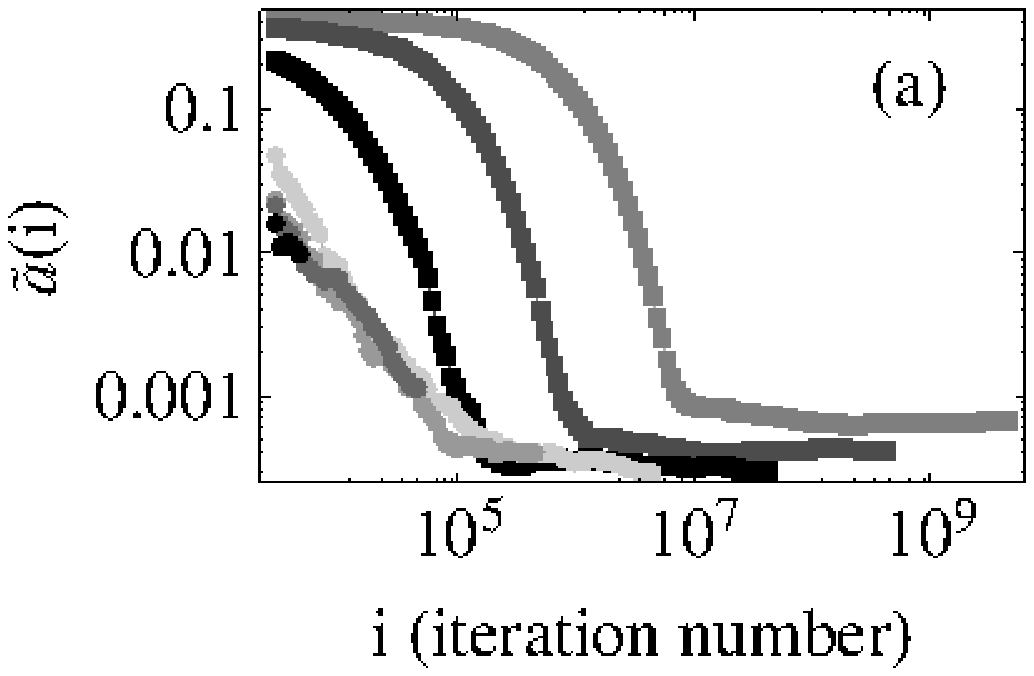} \\
\includegraphics[width=\xw]{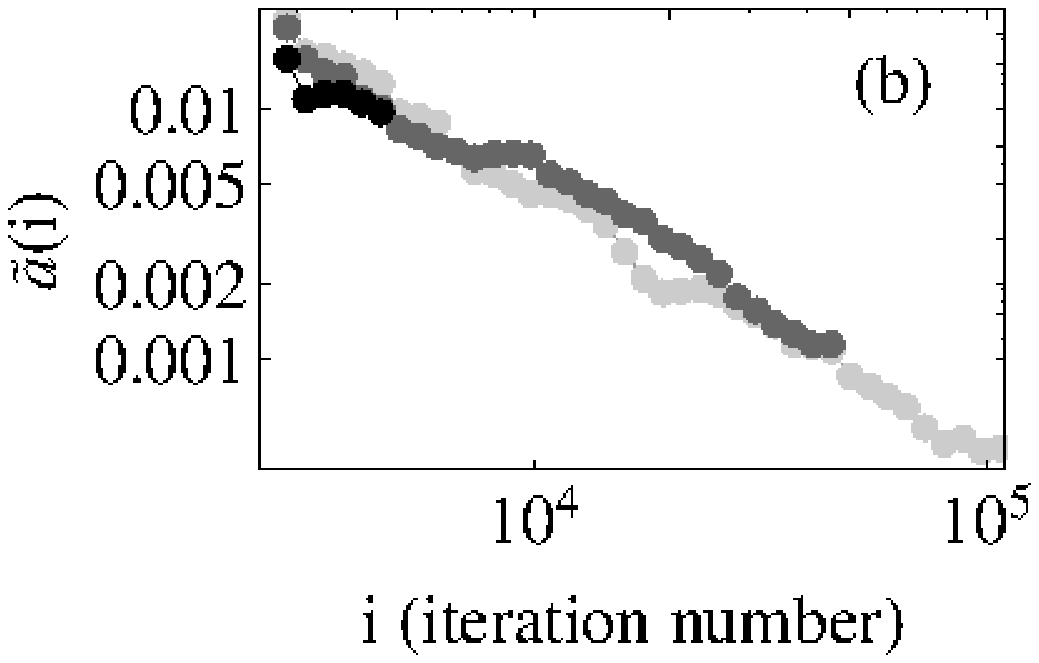} \\
\caption{(a) For the random $N=128$ instance, instantaneous acceptance rate vs
  iteration number $i$ for (from left to right) $I=10^4, 10^5 ,10^6, 10^7, 10^8, 10^9$ and  $10^{10}$.  (b) Detail for
  $I=10^4$ (black),  $10^5$ (medium gray),  $10^{6}$ (light gray).  }
\label{pir}
\end{figure}

\begin{figure}
\includegraphics[width=\xw]{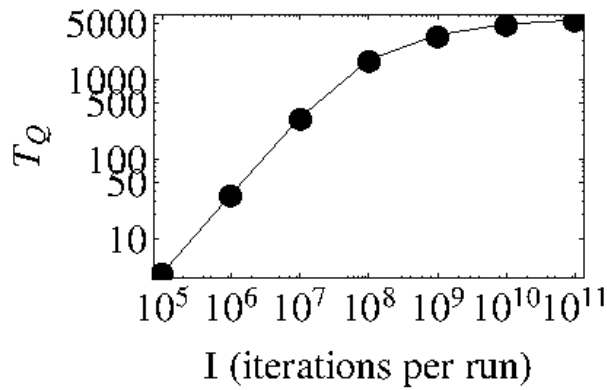}
\caption{For the random $N=128$  instance, temperature $T_Q$ versus number of iterations, I,  specified for run.  $T_Q$ is the temperature at which \
the specified number of consecutive iterations in which a swap is not performed is encountered.  The temperature is not lowered below this value. }
\label{ptx}
\end{figure}

To study the acceptance rate  $a(I)$ at a more granular level,  for various values of
$I$, Fig. \ref{pir} plots the instantaneous acceptance rate $\tilde a(i)$ versus the iteration number $i$ for $N=128$;
plots for other values of $N$ are similar.  For $I \gtrsim 10^6$, the
plots for each value of $I$ eventually reach constant
values $a_x(I)$.  The plots become constant at the iteration at which no swaps have been performed in the previous $Q$ iterations at which point the temperature is fixed at $T_Q$.  We also note that the constant values $a_x(I)$
increase with increasing $I$.  This behavior can be understood by
plotting $T_Q$.  For the random $N=128$ instance, in Fig. \ref{ptx} we
plot $T_Q$ as function of $I$ and\ note that $T_Q$ increases with
increasing $I$.  This reflects the fact that because of the the slower
cooling for larger number of iterations, SA finds a deeper local
minima from which it cannot escape easily at a higher temperatures for
larger $I$.  The increase in $a_x(I)$ as $I$ increases explains the increase in $a(I)$ for larger values of $I$  in  Fig. \ref{pacc} (a).

For $I \lesssim 10^6$, the plots in Fig. \ref{pir} terminate at lower and lower values as $I$ increases.   This explains the initial decrease in the acceptance rate as a function of $I$ in Fig. \ref{pacc}(a). 
The plots do not reach a constant value and terminate at a higher value than plots with $I
\gtrsim 10^6$.  This occurs because despite the low temperatures at
the end of run which inhibit cost-raising swaps, there are sufficient
numbers of cost-reducing swaps to maintain a relatively high value of
$\tilde a(i)$.

\subsection{Structured instances}
\label{accsApp}

\begin{figure}
\includegraphics[width=\xw]{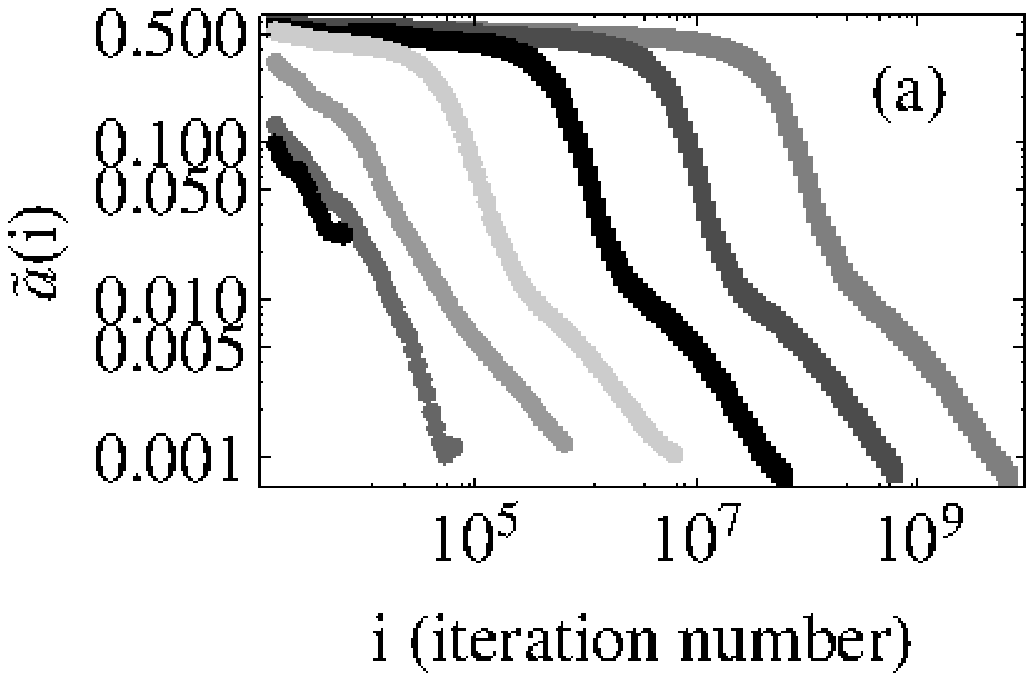} \\
\includegraphics[width=\xw]{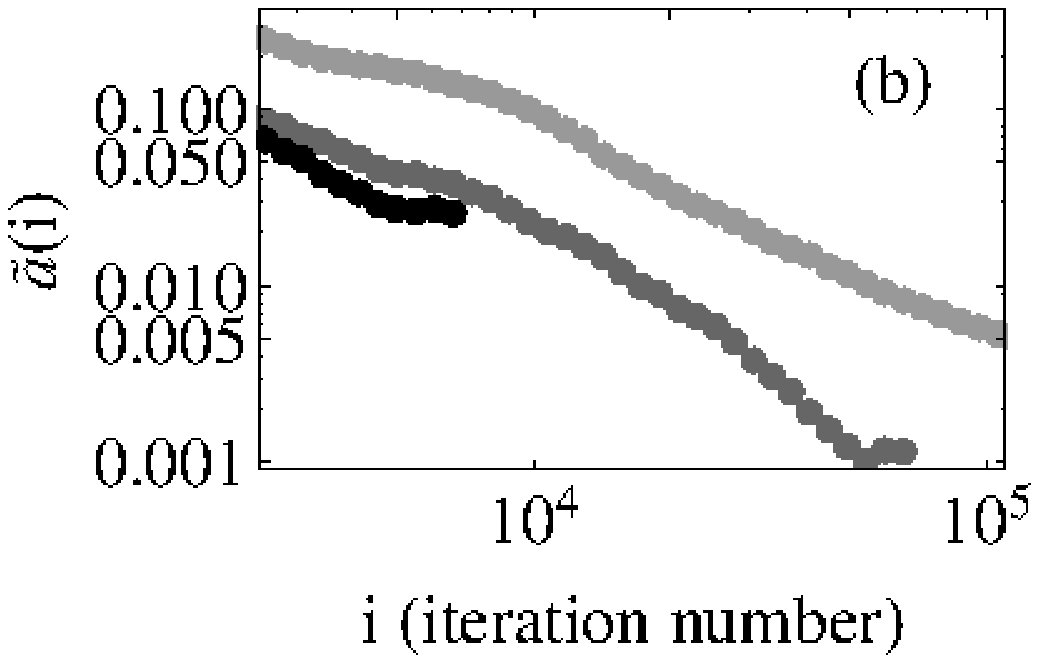} \\
\caption{(a) For the structured $N=125$ instance,  instantaneous acceptance rate vs
  iteration number $i$ for (from left to right)  $I=10^4, 10^5 ,10^6, 10^7, 10^8, 10^9$ and  $10^{10}$.  (b) Detail
  for $i=10^4, 10^5$ and $10^{6}$.  }
\label{pis}
\end{figure}

As with the random instances, in order to understand the behavior in
Fig. \ref{pacc}(b) we must study the instantaneous acceptance rate $\tilde a(i)$.   For various values of $I$, Fig. \ref{pis}
plots $\tilde a(i)$ versus the $i$.  As opposed to the random
instances, all plots decrease monotonically and never level off to
a constant value. This behavior is explained by the fact that for the
structured instances studied, the limit $Q$ on the number of
consecutive iterations in which no swap is made is never reached;
there are always low cost swaps which can be made.  The temperature
always decreases to $T_f$ and as the temperature decreases $\tilde a(i)$
decreases.  Because the plots in Fig. \ref{pis}(b) for larger $I$ all scale with the number of iterations, the value reached by the plots in Fig. \ref{pacc}(b) remain constant for large $I$.

Similar to the behavior for the random instances and for the same reason, for $I \lesssim
10^6$, the $\tilde a(i)$ plots terminate at a higher value than plots
with $I \gtrsim 10^6$.  As with the random instances, this explains the initial decrease in the acceptance rate as a function of $I$ in Fig. \ref{pacc}( b).

\section{Supplementary material }
 \label{supp}

The c++ program which implements the SA heuristic  described in this article can be found, in the online version, at doi:xxxxxx.

\section*{References}

\begin{hangref}

\item Anstreicher, K., 2003. Recent advances in the solution of quadratic assignment problems. Mathematical Programming 97, 27-42.

\item Applegate, D.L.  Bixby, R.E,   Chvatal, V.,  Cook, W.J., 2007.  The Traveling Salesman Problem: A Computational Study, Princeton University Press.

\item Burkard,  R.E., Cela, E.,  Karish,S. E., Rendl, F.,  1997.  QAPLIB - A Quadratic Assignment Problem Library.  J. Global Optim. 10 (1997) 391-403; http://www.seas.upenn.edu/qaplib/

\item Burkard, R. E., Dell'Amico, M. \& Martello, S.,  2009.   Assignment Problems, SIAM Philadelphia.

\item Burkard R. E.,  Rendl, F., 1984.  A thermodynamically motivated simulation procedure for combinatorial optimization problems.  EJOR 17, 169-174,

\item Cela E., 1998. The Quadratic Assignment Problem: Theory and Algorithms. Kluwer, Boston, MA.

\item Connolly, D. T.,  1990.  An improved annealing scheme for
    the QAP, EJOR 46 93-100.

\item de Carvalho Jr., S. A., Rahmann, S., 2006. Microarray
  layout as a quadratic assignment problem. In Hudson, D. {\it et al.},
  eds. {\it German Conference on Bioinformatics (GCB)}, {\it Lecture
    Notes in Infomatics\/}.  P-83, 11--20 .

\item Dickey, J., Hopkins, J.  1972. Campus building arrangement
  using TOPAZ. Transportation Res.  6, 59--68.

\item Drezner, Z., Hahn,P.M., Taillard \`E. D., 2005. Recent Advances for the quadratic assignment problem with special emphasis on instances that are difficult for meta-heuristic methods. Annals of Operations Research 139, 65-94.

\item  Elshafei, A. N,  1977.  Hospital layout as a quadratic
  assignment problem. Operations Res. Quarterly 28,  167--179.    

\item James, T., Rego, C., Glover, F., 2009a. Multistart Tabu Search  and Diversification Strategies for the Quadratic Assignment
  Problem. IEEE Tran. on Systems, Man, and Cybernetics  PART A: SYSTEMS AND HUMANS 39, 579-596.

\item  Kol\'a\v{r}, M., L\"assig, M., Berg, J.,  2008.  From protein
  interactions to functional annotation: Graph alignment in Herpes.    BMC Systems Biol. 2, Article 90.

\item Koopmans T., Beckmann, M., 1957. Assignment problems and the location of economic activities. Econometrica 25, 53-76.

\item Loiola,E.M., de Abreu, N.M.M., Boaventuro-Netto, P.O., Hahn, P., Querido, T.,  2007. A survey for the quadratic assignment problem. European Journal of Operational Research 176, 657-690.

\item Pardalos, P. M., Pitsoulis, L. S., Resende, M. G. C., 1997. Algorithm 769: Fortran subroutines for approximate solution of sparse quadratic assignment problems using GRASP. ACM Transactions on Mathematical Software (TOMS) 23, 196-208.

\item Paul, G., 2010. Comparative performance of tabu search and simulated annealing heuristics for
the quadratic assignment problem. Operations Research Letters 38 577-581

\item Steinberg, L., 1961. The backboard wiring problem: A  placement algorithm.  SIAM Rev.  3, 37-50.

\item Taillard, E., 1991.  Robust taboo search for the quadratic
  assignment problem. Parallel Computing 17, 443-455; code available
  at:http://mistic.heig-vd.ch/taillard/codes.dir/tabou\_qap.cpp.

\end{hangref}
\end{document}